\documentclass[11pt]{article}

\usepackage[final]{acl}
\usepackage{multirow}
\usepackage[table]{xcolor}
\usepackage{times}
\usepackage{latexsym}
\usepackage{array}
\usepackage{booktabs}
\usepackage{amsmath,amssymb}
\usepackage[T1]{fontenc}
\usepackage{booktabs}
\usepackage{multirow}
\usepackage{array}
\usepackage{makecell}
\definecolor{softgray}{gray}{0.93}
\usepackage{tabularx}
\colorlet{rowgray}{black!10}
\usepackage{pgfplots}
\pgfplotsset{compat=1.18}
\usetikzlibrary{patterns}
\usepgfplotslibrary{groupplots}
\usepackage{subcaption}
\usepackage[most]{tcolorbox}
\newcommand{\dropscore}[1]{\textcolor{blue}{#1}}
\definecolor{attnBlue}{RGB}{31, 119, 180}
\definecolor{attnGreen}{RGB}{44, 160, 44}
\definecolor{attnOrange}{RGB}{255, 127, 14}
\definecolor{attnRed}{RGB}{214, 39, 40}

\usepackage[utf8]{inputenc}

\usepackage{microtype}

\usepackage{inconsolata}

\usepackage{graphicx}
\usepackage{pgfplots}
\pgfplotsset{compat=1.18}
\usepgfplotslibrary{groupplots}
\title{SelFusion: Self-distillation for Diffusion Language Models}

\author{
  Hyeong Soo Lim\textsuperscript{1,*} \quad Jin Young Kim\textsuperscript{1,*} \quad Eun Seo Seo\textsuperscript{1} \quad Min Ho Jang\textsuperscript{1} \quad Ji Won Yoon\textsuperscript{1,\dag} \\
  \textsuperscript{1}Department of Artificial Intelligence, Chung-Ang University \\
  \texttt{\{andrew1001, wlsdud338, jeo0534, sunbi8534, jiwonyoon\}@cau.ac.kr} \\
  \textsuperscript{*}Equal contribution \quad \textsuperscript{\dag}Corresponding author
}

\begin{document}
\maketitle
\begin{abstract}

Diffusion language models (DLMs) alleviate the inherent latency bottleneck of autoregressive (AR) large language models (LLMs), but their degraded generation quality limits practical applicability.
Although knowledge distillation (KD) can be a promising direction for improving performance, we empirically find that naively applying conventional KD yields only marginal gains, or even degrades generation quality.
Based on these observations, we propose a novel self-distillation framework for DLMs, namely SelFusion.
To enable effective KD without an external teacher model, SelFusion performs two forward passes with different masking levels, defining the hard mode with a larger masking probability and the easy mode with a smaller masking probability.
However, the easy mode is not always more accurate than the hard mode and can be overconfident on incorrect tokens.
Thus, we introduce bidirectional KD between the two modes, which can dynamically determine the distillation direction based on token-level correctness.
Experimental results on instruction-following tasks show that the proposed self-distillation substantially outperforms other KD methods with external LLM and DLM teachers. In many configurations, the student trained with SelFusion even surpasses the performance of the LLM teacher, providing a practical path toward improving DLM generation quality. Source
code can be found at \url{https://github.com/scai-research/SelFusion_official}
\end{abstract}

\section{Introduction}

Recently, diffusion language models (DLMs) have emerged as a compelling alternative autoregressive (AR) large language models (LLMs). By leveraging parallel decoding mechanisms, DLMs offer faster inference capabilities, making them highly suitable for real-time applications. LLaDA \citep{LLaDA} and SMDM\citep{nie2025scalingmaskeddiffusionmodels} have demonstrated notable inference speedups over AR counterparts. However, DLMs typically underperform LLMs in terms of generation quality due to their non-autoregressive (NAR) nature  \citep{LLaDA,nie2025scalingmaskeddiffusionmodels,MDLM}.

\begin{figure}[t]
    \centering
    \includegraphics[width=\linewidth]{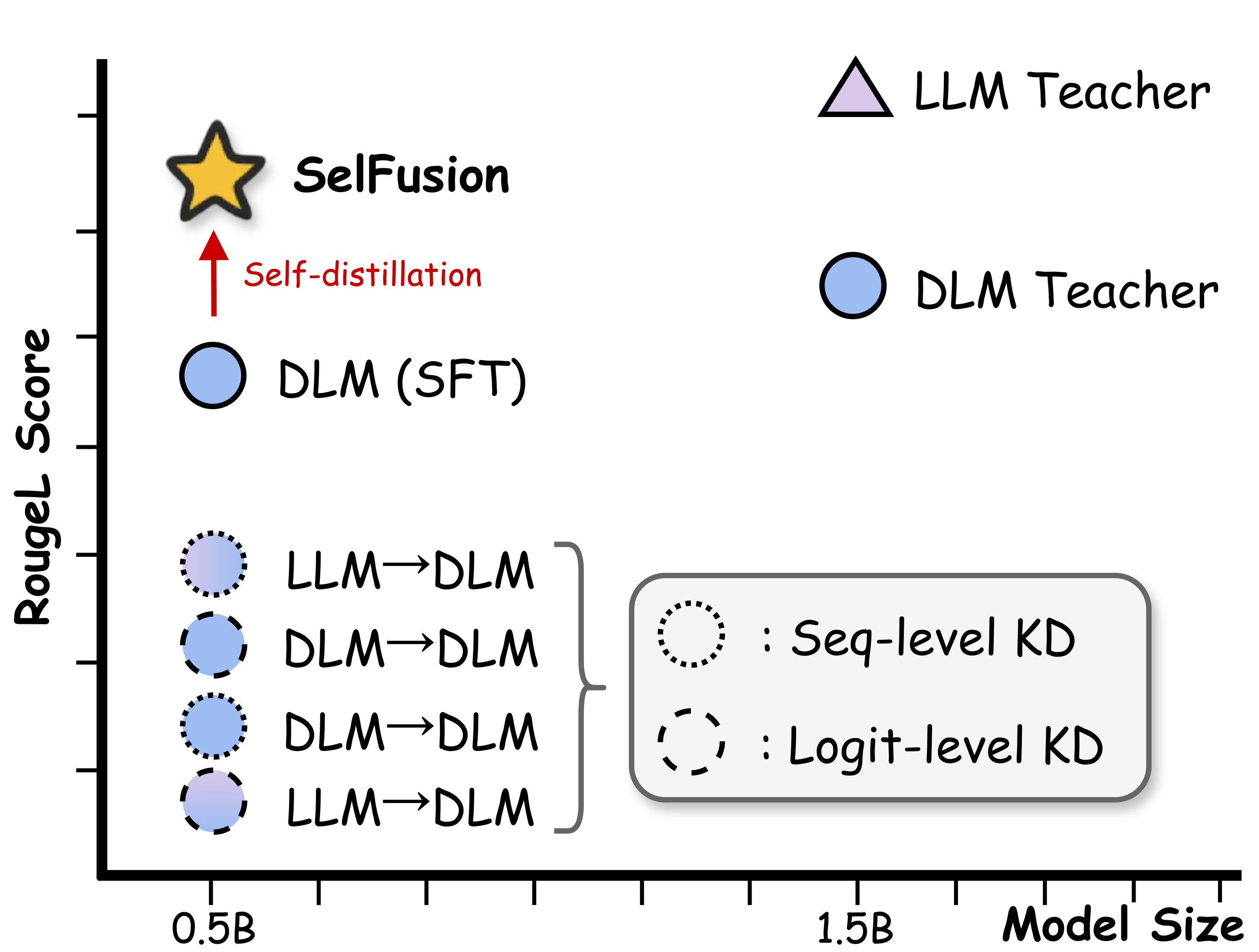}
    \caption{Rouge-L scores on the Dolly dataset. Existing KD methods for DLMs often underperform the SFT baseline, regardless of whether the teacher is DLM or LLM. In contrast, SelFusion significantly outperforms all baselines at the same model size.}
    \label{fig:teaser}
\end{figure}

To bridge this performance gap, knowledge distillation (KD) \citep{hinton2015distillingknowledgeneuralnetwork} can be a promising direction, enabling a student to mimic the behaviors of a strong teacher. In the context of LLMs, KD has been extensively studied and is typically categorized into two strategies.
First, logit-level KD performs distribution matching between teacher and student, which is the most common approach \citep{chen2024proxykd,minillm}. Second, sequence-level KD trains the student on teacher-generated text,  remaining useful when teacher distributions are inaccessible \citep{kim2016sequencelevelknowledgedistillation}.
While these strategies have proven effective in improving LLMs, their extension to the DLMs remains largely underexplored.

To examine the applicability of KD to DLMs, we distill DLM students from both LLM and DLM teachers. Surprisingly, as shown in Figure~\ref{fig:teaser}, distilled students achieve only marginal gains, or even exhibit performance degradation. In the case of LLM-to-DLM distillation, the distribution mismatch between the AR teacher and the NAR student limits the effectiveness of logit-level knowledge transfer, which will be further discussed in Section~\ref{sec:motivation}.
Sequence-level KD relies solely on teacher-generated outputs and thus provides improvements over logit-level supervision, but remains limited in terms of performance gains.
Moreover, DLM-to-DLM KD scenarios remain suboptimal, largely due to the limited generation quality of the DLM teacher.

Motivated by these observations, we propose \textbf{SelFusion}, a novel self-distillation framework for DLMs. Specifically, SelFusion leverages the noising process of DLMs to enable two forward modes within a single model, namely the easy mode and the hard mode. 
The input to the easy mode has a lower masking ratio than that of the hard mode. Since fewer tokens are masked, the easy mode is expected to yield more accurate predictions and provide more beneficial knowledge for KD.
However, it is not always more accurate than the hard mode and can also be overconfident on incorrect tokens. Thus, we introduce bidirectional KD between the easy and hard modes, dynamically determining the distillation direction by evaluating token-level correctness.

We evaluate SelFusion on multiple instruction-following benchmarks against existing KD methods. 
The proposed self-distillation consistently outperforms conventional KD baselines that depend on external teacher models across all configurations.
More surprisingly, SelFusion even surpasses the teacher models, including both DLMs and LLMs.
These results suggest that self-distillation with two modes effectively transfers knowledge within a single model.



\section{Related Work}
\label{sec:related_work}
\subsection{Diffusion Language Models}
DLMs for text generation can be categorized into continuous and discrete methods. Continuous methods embed tokens into continuous space, while discrete methods operate directly on token space \citep{gulrajani2023plaidlikelihoodbased}. Recently, masked diffusion has been predominantly adopted among discrete approaches, where the forward process progressively masks tokens and the reverse process learns to predict them \citep{MDLM, lou2024discrete,nie2025scalingmaskeddiffusionmodels, LLaDA}. Generally, increasing the number of denoising steps improves generation quality. LLaDA scaled masked diffusion to 8B parameters, demonstrating the potential of DLMs for fast inference through parallel generation \citep{LLaDA}. Despite these advances, DLMs still lag behind AR models in generation quality. For instance, recent DLMs underperform AR baselines by 10-32\% in perplexity \citep{LLaDA}. This gap highlights the need for further improvements in DLM generation quality. 

\subsection{Knowledge Distillation for Language Models}
KD \citep{romero2015fitnetshintsdeepnets,shridhar-etal-2023-distilling-scot,hsieh-etal-2023-distillingstepbystep, li2024promptkdunsupervisedpromptdistillation,jung2025todi} is a promising approach to improve model performance by transferring knowledge from a teacher model. Early work in AR models proposed logit-level KD that matches output logits \citep{hinton2015distillingknowledgeneuralnetwork}, followed by sequence-level KD that trains on teacher-generated sequences \citep{kim2016sequencelevelknowledgedistillation}. Recent advances have improved distillation for generative models. MiniLLM \citep{minillm} addressed the limitations of forward KL divergence by proposing reverse KL divergence with on-policy optimization for instruction-following tasks, while GKD \citep{on-policy_distillation} explored on-policy distillation using student-generated samples. Self-distillation methods have also shown promise by training models to match their own predictions from different configurations \citep{self_distillation, emnetwork-v202-yoon23a,self-acl-yang-etal-2024-self}. 
However, KD for DLMs to improve generation quality remains largely unexplored. Furthermore, existing work on DLM distillation has primarily focused on the pretraining phase, leaving post-training distillation scenarios unaddressed.

\begin{figure*}[t]
    \centering
    \includegraphics[width=1.0\textwidth]{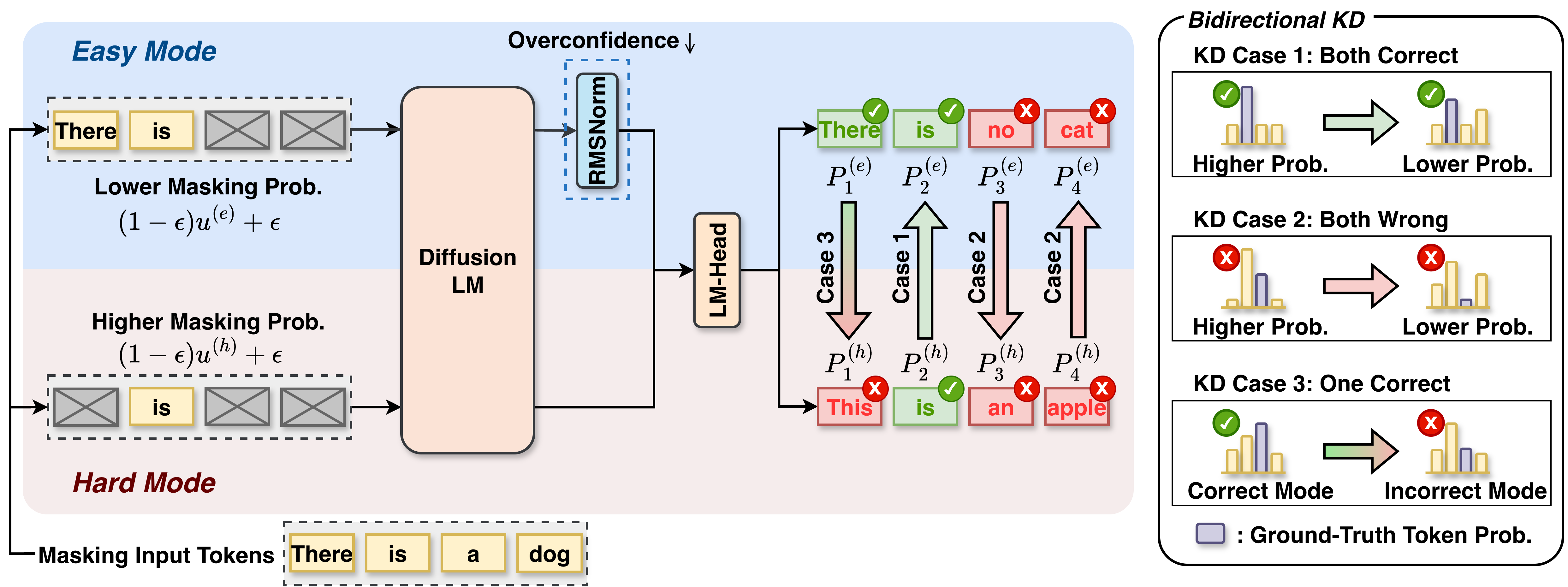}
    \caption{Overall architecture of SelFusion.
The framework comprises easy and hard modes that generate tokens under less and more masked context, respectively. Bidirectional KD determines the distillation direction based on token-level correctness.
}
    \label{fig:method}
\end{figure*}

\section{Methodology}
\subsection{Motivation}
\label{sec:motivation}

\paragraph{LLM-to-DLM Distillation.}
As aforementioned, logit-level distillation from an AR teacher is often ineffective due to distribution mismatch. We empirically observe that AR models assign near 100\% probabilty to the top-1 token, whereas DLMs exhibit approximately 60\% probability at a 50\% masking ratio. This gap hinders effective knowledge transfer, as the student struggles to match the teacher's spiky predictions. We also provide a detailed analysis of this mismatch in Section~\ref{sec:distribution_mismatch}. 

\paragraph{DLM-to-DLM Distillation.}
In the DLM-to-DLM setting, distillation is limited by the absence of sufficiently high-quality DLM teachers.
Even when DLM teachers are available, their generation quality remains substantially lower than that of AR models. As a result, logit-level KD yields only marginal gains relative to its increased training cost. Sequence-level KD also shows limited effectiveness, even when the teacher generates outputs with more denoising steps. 

\subsection{SelFusion}
Based on our findings that DLMs lack an effective teacher for KD, we propose a novel self-distillation, namely SelFusion. 
The overall process of SelFusion is illustrated in Figure \ref{fig:method}.

\paragraph{Two Modes with Different Masking.} 

The key idea is to leverage the noising process of DLMs to construct `easy mode' and `hard mode' within the same model. 
Specifically, while the hard mode follows the original random masking scheme, the easy mode is designed to use a lower masking ratio to expose more context, resulting in relatively higher masking ratios for the hard mode.
As a result, the easy mode is expected to yield more accurate predictions.
For example, in our experiments, the easy mode assigns approximately 10\% higher probability to the correct token than the hard mode throughout training.
This behavior is consistent with prior KD studies that emphasize student-friendly teachers, which maintain output distributions close to the student \citep{minillm, kim2024promptkddistillingstudentfriendlyknowledge, lee2024mentorkdmakingsmalllanguage}.
The easy mode tends to serve as the teacher for the hard mode, as it applies less masking and can thus provide relatively more accurate knowledge.
Further details are provided in Section~\ref{sec:easy_hard_reliability}.

\paragraph{Bidirectional KD.} 
\label{sec:bidirctional_kd}
However, the easy mode with lower masking does not always guarantee correct predictions, which motivates us to adaptively determine the distillation target. Therefore, we additionally present bidirectional KD, where the distillation direction for each token is determined based on correctness and confidence, considering three cases:
\begin{itemize}
    \item Both correct: When both modes predict correctly, the mode assigning higher probability to the predicted token serves as the distillation target.
    \item Both wrong: When both modes predict incorrectly, the mode assigning higher probability to the ground-truth token serves as the distillation target.
    \item One correct: When only one mode predicts correctly, that mode serves as the distillation target.
\end{itemize}
Formally, the token-level distillation direction $\mathcal{D}_t$ is defined as follows:
\begin{equation}
\mathcal{D}_t = \left\{
\begin{aligned}
&e \rightarrow h, && \text{if } c_e > c_h, \\
&h \rightarrow e, && \text{if } c_h > c_e, \\
&\arg\max_{m \in \{h, e\}} p_m(y_e \mid x), && \text{if } c_h = c_e.
\end{aligned}
\right.
\end{equation}
where $c_h$ and $c_e$ are binary indicators of correctness for the hard and easy mode predictions, respectively.
Distillation therefore follows the more reliable prediction at the token level.

\begin{figure}[t]
    \raggedright
    \begin{tikzpicture}
        \begin{axis}[
            width=\linewidth,
            height=6.0cm,
            xlabel={Token Rank},
            xlabel shift = -5.5pt, 
            ylabel={Prediction Probability (\%)}, 
            ylabel shift = -5.5pt, 
            xmin=1, xmax=5,
            ymin=0, ymax=105, 
            xtick={1,2,3,4,5},
            ytick={0, 20, 40, 60, 80, 100}, 
            ymajorgrids=true,
            grid style=dashed,
            legend style={
                at={(0.98,0.98)},
                anchor=north east,
                font=\small,
                draw=none, fill=none,
                cells={anchor=west}
            },
            label style={font=\normalsize},
            tick label style={font=\normalsize}
        ]
        \addplot[color=blue, mark=*, mark size=2.5pt, line width=0.8pt]
            coordinates {
            (1,99.95)(2,49.99)(3,33.33)(4,25.00)(5,20.00)
            };
            \addlegendentry{LLM (AR)}

        \addplot[color=orange, mark=triangle*, mark size=3.5pt]
            coordinates {
            (1,81.37)(2,43.80)(3,29.97)(4,22.76)(5,18.36)
            };
            \addlegendentry{DLM (MR=0.25)}

        \addplot[color=teal, mark=diamond*, mark size=4pt]
            coordinates {
            (1,70.43)(2,39.04)(3,26.95)(4,20.61)(5,16.73)
            };
            \addlegendentry{DLM (MR=0.50)}

        \addplot[color=purple, mark=pentagon*, mark size=3pt]
            coordinates {
            (1,42.53)(2,25.93)(3,18.89)(4,14.87)(5,12.35)
            };
            \addlegendentry{DLM (MR=0.75)}

        \end{axis}
    \end{tikzpicture}
    \caption{Comparison of average prediction probabilities for masked tokens in the Dolly dataset for LLM and DLM. For a fair comparison, we compute token probabilities using the same token prediction procedure as in training. Unlike AR models, DLMs exhibit flatter distributions, especially at higher MR.}
    \label{fig:maksing-level}
\end{figure}
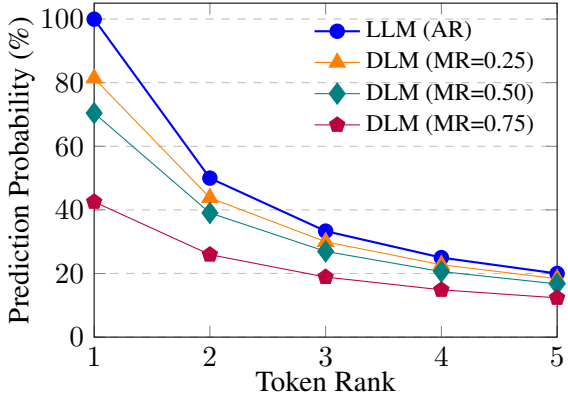

\paragraph{RMSNorm-based Logit Calibration.}
\label{sec:RMSNorm-method}
Since the easy mode has access to more context during token generation, it tends to produce overconfident predictions. Figure~\ref{fig:maksing-level} shows that lower masking ratios lead to more peaked distributions. More importantly, this overconfidence occurs not only on correct tokens but also on incorrect ones. Given that both modes predict incorrectly in approximately 40\% of cases, such overconfidence from the easy mode can hinder effective distillation. To address this, we apply RMSNorm-based logit calibration. Specifically, we insert RMSNorm between the final hidden state and the LM head of the easy mode. This selectively suppresses overconfidence, preserving confidence for correct predictions while substantially reducing it for incorrect ones. With only 1,280 parameters, RMSNorm alleviates overconfident predictions, enabling more effective knowledge transfer. Further analysis of RMSNorm is provided in Section~\ref{sec:RMSNorm}.

\subsection{Objective Function for SelFusion}
\label{sec:obective}
SelFusion jointly optimizes the hard mode diffusion loss, the easy mode diffusion loss, and the token-wise bidirectional distillation loss. 
Let $x$ be an input sequence and $y_i$ the ground-truth token at position $i$.

\paragraph{Masking Process and Notation.}
For each mode $m\in\{e,h\}$, we sample a noise level $u^{(m)}\in(0,1)$ and define the per-position masking probability as follows:
\begin{equation}
q_i^{(m)} \;=\; (1-\epsilon)\,u^{(m)} + \epsilon,
\end{equation}
where $\epsilon$ is a small constant. We then sample a binary mask variable $z_i^{(m)}\sim\mathrm{Bernoulli}(q_i^{(m)})$ independently for each position $i$. If $z_i^{(m)}=1$, the token at position $i$ is replaced by the special \textsc{[mask]} token; otherwise it remains visible.
We denote by $\mathcal{M}_m=\{i \mid z_i^{(m)}=1\}$ the set of masked positions in mode $m$.
The model output distribution at position $i$ in mode $m$ is denoted by $P_i^{(m)}(\cdot)$.

\paragraph{Diffusion Losses for Easy and Hard Mode.}
We compute the diffusion token-prediction losses over the masked positions, reweighted by the corresponding masking probabilities, which are defined as
\begin{align}
\mathcal{L}_{\mathrm{diff}}^{(h)}
&=
\frac{1}{|\mathcal{M}_h|}
\sum_{i\in\mathcal{M}_h}
\frac{-\log P_i^{(h)}(y_i)}{q_i^{(h)}},
\\
\mathcal{L}_{\mathrm{diff}}^{(e)}
&=
\frac{1}{|\mathcal{M}_e|}
\sum_{i\in\mathcal{M}_e}
\frac{-\log P_i^{(e)}(y_i)}{q_i^{(e)}}.
\end{align}

\begin{table*}[t]
\centering
{\fontsize{9}{12}\selectfont
\renewcommand{\arraystretch}{1.3}
\begin{tabular}{
>{\centering\arraybackslash}m{2.2cm}|
>{\centering\arraybackslash}m{3.0cm}|
>{\centering\arraybackslash}m{0.9cm}|
>{\centering\arraybackslash}m{1.25cm}|
>{\centering\arraybackslash}m{1.25cm}|
>{\centering\arraybackslash}m{1.25cm}|
>{\centering\arraybackslash}m{1.25cm}|
>{\centering\arraybackslash}m{1.25cm}
}
\noalign{\hrule height 1.5pt}
\textbf{Method} & \textbf{Model} & \textbf{Size} &
\cellcolor{softgray}\textbf{Dolly} &
\cellcolor{softgray}\textbf{Self-inst} &
\cellcolor{softgray}\textbf{Vicuna} &
\cellcolor{softgray}\textbf{Sinst} &
\cellcolor{softgray}\textbf{Uinst} \\
\noalign{\hrule height 1.5pt}
\multirow{6}{*}{SFT}
& $\text{LLM}_{tea}$ & \multirow{3}{*}{1476M}
& 24.4765 & 11.0382 & 14.9436 & 23.2118 & 27.1273 \\
\cline{2-2}\cline{4-8}
& $\text{DLM}_{tea}$ (8 Steps) &
& 17.2631 & 9.7284 & 12.4577 & 22.6308 & 21.9190 \\
\cline{2-2}\cline{4-8}
& $\text{DLM}_{tea}$ (16 Steps)&
& 18.6108 & 10.5172 & 14.7385  & 24.3427 & 23.9995 \\
\cline{2-8}
& $\text{LLM}_{stu}$ & \multirow{3}{*}{472M}
& 25.5660 & 11.1115 & 14.6017 & 22.4033 & 25.2105 \\
\cline{2-2}\cline{4-8}
& $\text{DLM}_{stu}$ (8 Steps)  &
&18.7688  & 10.6877 & 15.0473 & 24.7139 & 23.8430 \\
\cline{2-2}\cline{4-8}
& $\text{DLM}_{stu}$ (16 Steps) &
& 19.5204 & 11.6416  & 16.3351 & 25.7185 & 25.5245 \\
\noalign{\hrule height 1.5pt}
\textbf{Method} & \textbf{KD} & \textbf{Steps} &
\cellcolor{softgray}\textbf{Dolly} &
\cellcolor{softgray}\textbf{Self-inst} &
\cellcolor{softgray}\textbf{Vicuna} &
\cellcolor{softgray}\textbf{Sinst} &
\cellcolor{softgray}\textbf{Uinst} \\
\noalign{\hrule height 1.5pt}
\multirow{2}{*}{Logit-level KD}
& $\text{LLM}_{tea}$$\rightarrow$$\text{DLM}_{stu}$ & \multirow{5}{*}{8} 
&15.1075  & 8.7333 & 14.3417 & 16.5709 & 17.8998 \\
\cline{2-2}\cline{4-8}  
& $\text{DLM}_{tea}$$\rightarrow$$\text{DLM}_{stu}$ &
& 17.2276 & 9.8326 & 14.4729 &20.4459  & 19.8666 \\
\cline{1-2}\cline{4-8} 
\multirow{2}{*}{Seq-level KD}
& $\text{LLM}_{tea}$$\rightarrow$$\text{DLM}_{stu}$ &
& 18.8576 &10.0266  & 14.3811 & 22.4884 & 22.6125 \\
\cline{2-2}\cline{4-8}
& $\text{DLM}_{tea}$$\rightarrow$$\text{DLM}_{stu}$ &
& 14.6822 & 8.3684 & 14.6454 & 16.2587 & 17.1783  \\
\cline{1-2}\cline{4-8}
\textbf{Ours, SelFusion}
& $\text{DLM}_{stu}$$\leftrightarrow$$\text{DLM}_{stu}$ &
& \textbf{21.3926} & \textbf{12.0022} & \textbf{16.6586} & \textbf{26.3464} & \textbf{26.4189} \\
\noalign{\hrule height 1pt}
\multirow{2}{*}{Logit-level KD}
& $\text{LLM}_{tea}$$\rightarrow$$\text{DLM}_{stu}$ & \multirow{5}{*}{16} 
& 15.3833 & 9.0501 & 15.4732 & 16.5188 & 17.9967 \\
\cline{2-2}\cline{4-8}  
& $\text{DLM}_{tea}$$\rightarrow$$\text{DLM}_{stu}$ &
& 17.6549 & 10.1753 & 16.0814 & 21.4601 & 21.4262 \\
\cline{1-2}\cline{4-8} 
\multirow{2}{*}{Seq-level KD}
& $\text{LLM}_{tea}$$\rightarrow$$\text{DLM}_{stu}$ &
 & 19.9478&11.0604  & 16.1340 & 23.7022 & 24.4449 \\
\cline{2-2}\cline{4-8}
& $\text{DLM}_{tea}$$\rightarrow$$\text{DLM}_{stu}$ &
&15.5604  &9.5634  & 16.4462 & 16.9650 & 18.4398 \\
\cline{1-2}\cline{4-8}
\textbf{Ours, SelFusion}
& $\text{DLM}_{stu}$$\leftrightarrow$$\text{DLM}_{stu}$ &
& \textbf{21.6317} & \textbf{12.8707} & \textbf{17.0788} & \textbf{27.0745} & \textbf{27.8595} \\
\noalign{\hrule height 1.5pt}
\end{tabular}
}
\caption{Performance comparison on multiple evaluation datasets. The first block reports SFT results of the teacher and student baselines, where $\mathrm{LLM}_{\mathrm{tea}}$ and $\mathrm{DLM}_{\mathrm{tea}}$ denote the teacher models and $\mathrm{LLM}_{\mathrm{stu}}$ and $\mathrm{DLM}_{\mathrm{stu}}$ denote the student baseline models. The second block reports KD results, where the KD column indicates the distillation direction. Bold indicates the best result.}
\label{tab:performance_comparison}
\end{table*}

\paragraph{Bidirectional KD Loss.}
Let $\mathcal{D}$ be the set of distillation positions.
For each $i\in\mathcal{D}$, the distillation direction $\mathcal{D}_i\in\{e\!\rightarrow\!h,\;h\!\rightarrow\!e\}$ is determined by the rule defined in Eq.~(1).
With temperature $T$, we define the temperature-scaled distributions from the logits $z_i^{(m)}$ as follows:
\begin{equation}
P^{(m)}_{i,T}(\cdot) \;=\; \mathrm{softmax}\!\left(z_i^{(m)}/T\right),
\end{equation}
where $\mathcal{V}$ denotes the vocabulary (with size $|\mathcal{V}|$), where $z_i^{(m)}\in\mathbb{R}^{|\mathcal{V}|}$ denotes the logits at position $i$ under mode $m\in\{h,e\}$.
We use the Kullback-Leibler (KL) divergence to measure the discrepancy between two output distributions. The bidirectional KD loss is given by
\begin{equation}
\mathcal{L}_{\mathrm{bkd}} = \frac{T^2}{|\mathcal{D}|}\sum_{i\in\mathcal{D}} \mathrm{KL}(P_{i,T}^{(j)} \| P_{i,T}^{(k)}),
\end{equation} 
where $(j, k) \in \{(e, h), (h, e)\}$ depending on the direction $\mathcal{D}_i$.

\paragraph{Total Objective.}
The final training objective can be calculated as
\begin{equation}
\mathcal{L}_{\mathrm{SelFusion}}
=
\mathcal{L}_{\mathrm{diff}}^{(h)} + \mathcal{L}_{\mathrm{diff}}^{(e)} + \mathcal{L}_{\mathrm{bkd}}.
\end{equation}
Since both modes share the same parameters, the combined loss updates both modes simultaneously in a single backward pass.

\section{Experiments}
\subsection{Experimental Settings}
\paragraph{Datasets.}
Following previous studies \citep{minillm, kim2024promptkddistillingstudentfriendlyknowledge}, we evaluated on instruction-following tasks, where the model generates responses conditioned on instructions. We used Databricks-Dolly-15K \citep{DatabricksBlog2023DollyV2} as our training dataset, with 12K samples for training and 500 samples for evaluation. We evaluated on five instruction-following benchmarks, including Dolly (500 samples), Self-Inst (252 samples) \citep{wang2023selfinstruct}, Vicuna (80 samples) \citep{vicuna2023}, and the $[11, +\infty)$ response-length subsets of S-NI (1,694 samples) \citep{wang2022s-instructionsgeneralizationdeclarativeinstructions} and UnNI (23,916 samples) \citep{honovich-etal-2023-uinst}. The five benchmarks described above are the evaluation datasets reported in Table~\ref{tab:performance_comparison}.

\paragraph{Models and Training Setup.}
All experiments were conducted using SMDM architectures with 472M and 1476M parameters~\citep{nie2025scalingmaskeddiffusionmodels}, with the number of training epochs fixed to 20. To identify the optimal configuration for each model and method, we explored various learning rates and epochs. The LLM teacher used in our experiments was a TinyLlama model with 1,476M parameters. We used the DLM and LLM teachers from~\citep{nie2025scalingmaskeddiffusionmodels}, where both models were trained under the same setup with matched training data, model size, and training epochs. Comprehensive details regarding the hyperparameter search space and final configurations are provided in the Appendix~\ref{sec:appendix}. All models were evaluated using the checkpoint from the final training step. Experiments were executed on four NVIDIA H200 GPUs, each with 141GB of memory.

\paragraph{Evaluation Configurations.}
Following prior work on DLMs~\citep{nie2025scalingmaskeddiffusionmodels}, we fixed the classifier-free guidance (CFG) scale to 1.0 for all DLMs. For LLMs, the sampling temperature was set to 1.0 during evaluation. Generation quality was assessed using the ROUGE-L metric~\citep{lin-2004-rouge}, which is widely adopted for evaluating instruction-following text generation. We evaluated each benchmark using three different random seeds and report the average across the three runs. Additional implementation and training details are provided in the Appendix~\ref{sec:appendix}.

\begin{table*}[t]
\centering
{\fontsize{9}{12}\selectfont
\renewcommand{\arraystretch}{1.3}
\begin{tabular}{
>{\centering\arraybackslash}m{2.4cm}|
>{\centering\arraybackslash}m{3.0cm}|
>{\centering\arraybackslash}m{2.3cm}|
>{\centering\arraybackslash}m{2.3cm}|
>{\centering\arraybackslash}m{1.8cm}
}
\noalign{\hrule height 1.5pt}
\textbf{Method} & \textbf{KD} & \textbf{Student training} & \textbf{Teacher training} & \textbf{Total} \\
\noalign{\hrule height 1.5pt}
SFT & --- & $2.8 \times 10^{2}$ & --- & $2.8 \times 10^{2}$ \\
\hline
\multirow{2}{*}{Seq-level KD} 
& $\mathrm{LLM}_{\mathrm{tea}} \rightarrow \mathrm{DLM}_{\mathrm{stu}}$ 
& $2.8 \times 10^{2}$ & $7.5 \times 10^{2}$ & $1.03 \times 10^{3}$ \\
\cline{2-5}
& $\mathrm{DLM}_{\mathrm{tea}} \rightarrow \mathrm{DLM}_{\mathrm{stu}}$ 
& $2.8 \times 10^{2}$ & $7.5 \times 10^{2}$ & $1.03 \times 10^{3}$ \\
\hline
\multirow{2}{*}{Logit-level KD} 
& $\mathrm{DLM}_{\mathrm{tea}} \rightarrow \mathrm{DLM}_{\mathrm{stu}}$ 
& $5.4 \times 10^{2}$ & $7.5 \times 10^{2}$ & $1.29 \times 10^{3}$ \\
\cline{2-5}
& $\mathrm{LLM}_{\mathrm{tea}} \rightarrow \mathrm{DLM}_{\mathrm{stu}}$ 
& $5.4 \times 10^{2}$ & $7.5 \times 10^{2}$ & $1.29 \times 10^{3}$ \\
\hline
\textbf{Ours, SelFusion} & $\mathrm{DLM}_{\mathrm{stu}} \leftrightarrow \mathrm{DLM}_{\mathrm{stu}}$ & $5.6 \times 10^{2}$ & --- & $\mathbf{5.6 \times 10^{2}}$ \\
\noalign{\hrule height 1.5pt}
\end{tabular}
}
\caption{Training cost analysis measured in TFLOPs. Although SelFusion requires higher per-iteration computation than SFT, it eliminates teacher training and thus reduces total training computation compared to KD methods that rely on a separately trained teacher. All methods are compared under the same training setup with 485 iterations.}
\label{tab:training_cost}
\end{table*}

\subsection{Experimental Results}
Firstly, we evaluated conventional logit-level and sequence-level KD in both the LLM-to-DLM and DLM-to-DLM settings, as shown in Table~\ref{tab:performance_comparison}. 
For logit-level KD, we minimized the KL divergence between the teacher and student distributions. Since the student was a DLM, we applied the KD loss only to the masked tokens, following the DLM generation principle.
In the case of sequence-level KD, we trained the student with teacher-generated outputs, as described in Section~\ref{sec:related_work}. This approach required the teacher that could generate high-quality target sequences to provide effective supervision \citep{kim2016sequencelevelknowledgedistillation}.
Prior work showed that DLMs’ generation quality can be improved by increasing the number of diffusion steps~\citep{deschenaux2025beyondllm,nie2025scalingmaskeddiffusionmodels,chen2025dlmone}. Thus, in the DLM-to-DLM setting, we generated teacher pseudo-targets using 64-step inference and performed sequence-level KD on these sequences. We presented results for 8 and 16 diffusion steps in Table~\ref{tab:performance_comparison}. 

We began by evaluating LLM-to-DLM distillation with 8 diffusion steps. From the results, it is confirmed that logit-level KD with the LLM teacher led to substantial performance degradation. For example, on Dolly, the score dropped to 15.11, compared to 18.77 for the DLM SFT baseline. This trend was consistent across all benchmarks, indicating that direct logit matching from the LLM teacher to the DLM student was ineffective. Sequence-level KD also did not surpass the DLM SFT baseline on most benchmarks, with Dolly as the only exception. We next evaluated DLM-to-DLM distillation using the pretrained DLM teacher. Distillation did not improve over the SFT baseline. For example, on Uinst with 8 diffusion steps, the DLM SFT model achieved 23.84, whereas logit-level KD reached only 19.87. Sequence-level KD further exhibited performance drops across benchmarks. These results suggested that the DLM teacher’s generation quality was insufficient to provide beneficial knowledge at either the sequence or logit level.

In contrast, the proposed self-distillation method achieved substantial performance gains without relying on any external teacher model. Table~\ref{tab:performance_comparison} shows that SelFusion outperformed the strongest competing baseline, LLM-to-DLM sequence-level KD, by 2 to 4 points, corresponding to an approximate 16\% relative improvement. Compared with the DLM SFT baseline, it consistently improved performance by 1.5 to 3 points across all five benchmarks. Notably, SelFusion also surpassed the LLM teacher on multiple benchmarks. For example, SelFusion achieved 12.87 on Self-inst and 17.08 on Vicuna, exceeding the LLM teacher scores of 11.04 and 14.95, respectively. It also improved from 23.21 to 27.07 on Sinst and from 27.13 to 27.86 on Uinst. Overall, these results demonstrated that our self-distillation design provides a practical path for DLMs, requiring no external teacher and introducing only 1,280 additional parameters.


\begin{table*}[t]
\centering
{\fontsize{9}{12}\selectfont
\renewcommand{\arraystretch}{1.25}
\begin{tabular}{
>{\centering\arraybackslash}m{3.6cm}
| >{\centering\arraybackslash}m{0.9cm}
| >{\centering\arraybackslash}m{1.5cm}
| >{\centering\arraybackslash}m{1.5cm}
| >{\centering\arraybackslash}m{1.5cm}
| >{\centering\arraybackslash}m{1.5cm}
| >{\centering\arraybackslash}m{1.5cm}
}
\noalign{\hrule height 1.5pt}
\textbf{Method} & \textbf{Steps} & 
\cellcolor{softgray}\textbf{Dolly} & 
\cellcolor{softgray}\textbf{Self-inst} & 
\cellcolor{softgray}\textbf{Vicuna} & 
\cellcolor{softgray}\textbf{Sinst} & 
\cellcolor{softgray}\textbf{Uinst} \\
\noalign{\hrule height 1.5pt}
\multirow{2}{3.6cm}{\raggedright \textbf{SelFusion}} & 
\multirow{8}{*}{16} & 
\multirow{2}{*}{\textbf{21.6317}} & 
\multirow{2}{*}{\textbf{12.8707}} & 
\multirow{2}{*}{\textbf{17.0788}} & 
\multirow{2}{*}{\textbf{27.0745}} & 
\multirow{2}{*}{\textbf{27.8595}} \\
& & & & & & \\
\cline{1-1}\cline{3-7}
\multirow{2}{3.6cm}{\raggedleft\arraybackslash \small \shortstack[r]{w/o bidirectional KD\\(easy$\rightarrow$hard)}\hspace*{-0.2em}}  & 
& 13.6755 & 8.5558 & 10.6129 & 21.8953 & 20.9376 \\
& 
& \multicolumn{1}{>{\raggedleft\arraybackslash}m{1.5cm}|}{\hfill \dropscore{(-7.9562)}} 
& \multicolumn{1}{>{\raggedleft\arraybackslash}m{1.5cm}|}{\hfill \dropscore{(-4.3149)}}
& \multicolumn{1}{>{\raggedleft\arraybackslash}m{1.5cm}|}{\hfill \dropscore{(-6.4659)}}
& \multicolumn{1}{>{\raggedleft\arraybackslash}m{1.5cm}|}{\hfill \dropscore{(-5.1792)}}
& \multicolumn{1}{>{\raggedleft\arraybackslash}m{1.5cm}}{\hfill \dropscore{(-6.9219)}} \\
\cline{1-1}\cline{3-7}
\multirow{2}{3.6cm}{\raggedleft\arraybackslash \small \shortstack[r]{w/o bidirectional KD\\(hard$\rightarrow$easy)}\hspace*{-0.2em}}  & 
& 15.7107 & 9.6470 & 11.0940 & 23.9227 & 23.2298 \\
& 
& \multicolumn{1}{>{\raggedleft\arraybackslash}m{1.5cm}|}{\hfill \dropscore{(-5.9210)}} 
& \multicolumn{1}{>{\raggedleft\arraybackslash}m{1.5cm}|}{\hfill \dropscore{(-3.2237)}}
& \multicolumn{1}{>{\raggedleft\arraybackslash}m{1.5cm}|}{\hfill \dropscore{(-5.9848)}}
& \multicolumn{1}{>{\raggedleft\arraybackslash}m{1.5cm}|}{\hfill \dropscore{(-3.1518)}}
& \multicolumn{1}{>{\raggedleft\arraybackslash}m{1.5cm}}{\hfill \dropscore{(-4.6297)}} \\
\cline{1-1}\cline{3-7}
\multirow{2}{3.6cm}{\raggedleft\arraybackslash \small w/o RMSNorm\hspace*{-0.2em}} & 
& 17.3566 & 11.0430 & 16.4204 & 21.2283 & 21.9992 \\
& 
& \multicolumn{1}{>{\raggedleft\arraybackslash}m{1.5cm}|}{\hfill \dropscore{(-4.2751)}} 
& \multicolumn{1}{>{\raggedleft\arraybackslash}m{1.5cm}|}{\hfill \dropscore{(-1.8277)}}
& \multicolumn{1}{>{\raggedleft\arraybackslash}m{1.5cm}|}{\hfill \dropscore{(-0.6584)}}
& \multicolumn{1}{>{\raggedleft\arraybackslash}m{1.5cm}|}{\hfill \dropscore{(-5.8462)}}
& \multicolumn{1}{>{\raggedleft\arraybackslash}m{1.5cm}}{\hfill \dropscore{(-5.8603)}} \\
\noalign{\hrule height 1.5pt}
\end{tabular}
}
\caption{Ablation results of SelFusion. We perform ablation studies by individually removing bidirectional KD and RMSNorm. For bidirectional KD, we further examine each unidirectional variant (easy→hard and hard→easy) to isolate the contribution of each direction. Across all benchmarks, removing either component leads to consistent performance degradation, and neither unidirectional variant matches the bidirectional setting}
\label{tab:selfusion_ablation}
\end{table*}

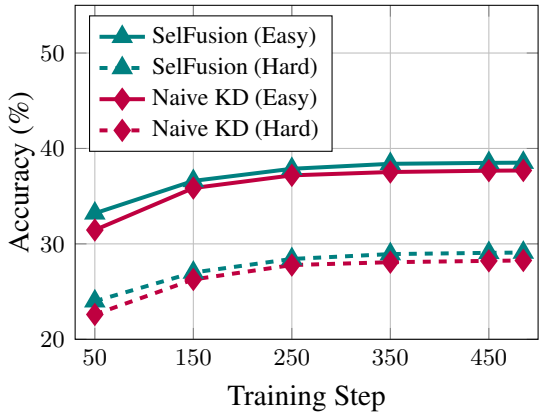
\begin{figure}[t]
    \raggedright
    \begin{tikzpicture}
        \begin{axis}[
            width=\columnwidth,
            height=6cm,
            xlabel={Training Step},
            ylabel={Accuracy (\%)},
            ylabel style={yshift=-0.5em},
            grid=major,
            legend pos=north west,
            legend cell align={left},
            legend style={font=\small},
            tick label style={font=\footnotesize},
            label style={font=\normalsize},
            ymin=20, ymax=55,
            xtick={50, 150, 250, 350, 450},
            xmin=30, xmax=500
        ]
        \addplot[
            color=teal,
            mark=triangle*,
            mark options={scale=1.6, solid, fill=teal},
            line width=1.5pt
        ] coordinates {
            (50,33.20)  (150,36.60)  (250,37.87)
            (350,38.39) (450,38.49) (485,38.51)
        };
        \addlegendentry{SelFusion (Easy)}
        
        \addplot[
            color=teal,
            mark=triangle*,
            mark options={scale=1.6, solid, fill=teal},
            dashed,
            line width=1.5pt
        ] coordinates {
            (50,23.99)  (150,27.00)  (250,28.42)
            (350,28.93) (450,29.06) (485,29.09)
        };
        \addlegendentry{SelFusion (Hard)}
        
        \addplot[
            color=purple,
            mark=diamond*,
            mark options={scale=1.6, solid, fill=purple},
            line width=1.5pt
        ] coordinates {
            (50,31.45)  (150,35.85)  (250,37.18)
            (350,37.53) (450,37.67) (485,37.69)
        };
        \addlegendentry{Naive KD (Easy)}
        
        \addplot[
            color=purple,
            mark=diamond*,
            mark options={scale=1.6, solid, fill=purple},
            dashed,
            line width=1.5pt
        ] coordinates {
            (50,22.60) (150,26.27)  (250,27.78)
            (350,28.08) (450,28.22) (485,28.26)
        };
        \addlegendentry{Naive KD (Hard)}

        \end{axis}
    \end{tikzpicture}
    \caption{
Training accuracy comparison between SelFusion and the naive KD baseline. Naive KD uses a fixed one-way distillation direction between the two modes, Easy$\rightarrow$Hard.
Accuracy is evaluated only on the masked positions for each of the Easy and Hard modes, which use different masking ratios.
}
    \label{fig:accuracy_comparison}
\end{figure}

\begin{figure}[t]
    \raggedright
    \begin{tikzpicture}
        \begin{axis}[
            width=1\linewidth,
            height=6.0cm,
            xlabel={Step},
            xlabel shift = -5.5pt, 
            ylabel={Mean probability (\%)},
            ylabel shift = -5.5pt, 
            ymin=35, ymax=63,
            ytick={35,40,45,50,55,60},
            ymajorgrids=true,
            grid style={dashed, line width=0.4pt, draw=gray!50},
            label style={font=\normalsize},
            tick label style={font=\small},
            symbolic x coords={
                50,sep1,sep2,sep3,
                250,sep4,sep5,sep6,
                450,sep7,sep8,sep9,
                485
            },
            xtick={50,250,450,485},
            ybar=3.5pt, 
            bar width=9pt, 
            enlarge x limits=0.15,
            nodes near coords,
            every node near coord/.append style={
                font=\small,
                yshift=1pt,
                xshift=-1pt,
                /pgf/number format/fixed,
                /pgf/number format/precision=3
            },
            legend style={
                font=\small,
                at={(0.02,0.98)},
                anchor=north west
            }
        ]

        \addplot[fill=teal!80, draw=teal] coordinates {
            (50,39) (250,42) (450,42) (485,42)
        };
        \addlegendentry{Hard (Student)}

        \addplot[fill=purple!80, draw=purple] coordinates {
            (50,45) (250,52) (450,53) (485,53)
        };
        \addlegendentry{Easy (Teacher)}

        \end{axis}
    \end{tikzpicture}
    \caption{Step-wise comparison of the mean probability assigned to the ground truth tokens by the easy and hard modes during training.}
    \label{fig:easy_hard_token_prob_by_step}
\end{figure}
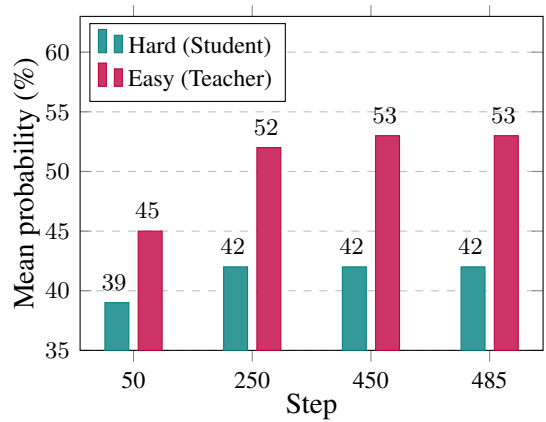

\begin{figure*}[t]
    \centering
    \begin{subfigure}[b]{0.495\textwidth}
        \centering
        \begin{tikzpicture}
            \begin{axis}[
                width=\linewidth,
                height=5.5cm,
                title={\textbf{(a) Both Correct}},
                title style={font=\normalsize},
                xlabel={Top-$k$ Rank},
                ylabel={Average Probability (\%)},
                xmin=0.5, xmax=5.5,
                ymin=0, ymax=105,
                xtick={1,2,3,4,5},
                ytick={0,20,40,60,80,100},
                grid=both,
                grid style={line width=0.3pt, draw=gray!30},
                major grid style={line width=0.5pt, draw=gray!50},
                label style={font=\normalsize},
                tick label style={font=\small},
                legend style={
                    at={(0.97,0.97)},
                    anchor=north east,
                    font=\footnotesize,
                    draw=gray!50,
                    rounded corners=2pt,
                    fill=white,
                    fill opacity=0.9,
                    cells={anchor=west},
                    row sep=2pt,
                },
                every axis plot/.append style={line width=1.2pt},
            ]
                \addplot[color=blue, mark=square*, mark size=3pt, mark options={solid, fill=blue}]
                    coordinates { (1,87.43)(2,4.43)(3,1.55)(4,0.86)(5,0.60) };
                \addlegendentry{w/o RMS (Easy,MR=50\%)}

                \addplot[color=blue, dashed, mark=*, mark size=3pt, mark options={solid, fill=blue}]
                    coordinates { (1,58.69)(2,4.07)(3,1.64)(4,1.06)(5,0.77) };
                \addlegendentry{with RMS (Easy,MR=50\%)}

                \addplot[color=red, mark=triangle*, mark size=3.5pt, mark options={solid, fill=red}]
                    coordinates { (1,76.76)(2,5.89)(3,2.46)(4,1.51)(5,1.08) };
                \addlegendentry{w/o RMS (Hard,MR=75\%)}
            \end{axis}
        \end{tikzpicture}
    \end{subfigure}
    \hfill
    \begin{subfigure}[b]{0.495\textwidth}
        \centering
        \begin{tikzpicture}
            \begin{axis}[
                width=\linewidth,
                height=5.5cm,
                title={\textbf{(b) Both Incorrect}},
                title style={font=\normalsize},
                xlabel={Top-$k$ Rank},
                ylabel={Average Probability (\%)},
                xmin=0.5, xmax=5.5,
                ymin=0, ymax=42,
                xtick={1,2,3,4,5},
                ytick={0,10,20,30,40},
                grid=both,
                grid style={line width=0.3pt, draw=gray!30},
                major grid style={line width=0.5pt, draw=gray!50},
                label style={font=\normalsize},
                tick label style={font=\small},
                legend style={
                    at={(0.97,0.97)},
                    anchor=north east,
                    font=\footnotesize,
                    draw=gray!50,
                    rounded corners=2pt,
                    fill=white,
                    fill opacity=0.9,
                    cells={anchor=west},
                    row sep=2pt,
                },
                every axis plot/.append style={line width=1.2pt},
            ]
                \addplot[color=blue, mark=square*, mark size=3pt, mark options={solid, fill=blue}]
                    coordinates { (1,37.08)(2,12.04)(3,6.30)(4,4.24)(5,3.15) };
                \addlegendentry{w/o RMS (Easy,MR=50\%)}

                \addplot[color=blue, dashed, mark=*, mark size=3pt, mark options={solid, fill=blue}]
                    coordinates { (1,10.34)(2,4.63)(3,2.83)(4,2.11)(5,1.71) };
                \addlegendentry{with RMS (Easy,MR=50\%)}

                \addplot[color=red, mark=triangle*, mark size=3.5pt, mark options={solid, fill=red}]
                    coordinates { (1,24.70)(2,9.54)(3,5.61)(4,4.02)(5,3.14) };
                \addlegendentry{w/o RMS (Hard,MR=75\%)}
            \end{axis}
        \end{tikzpicture}
    \end{subfigure}%
    \caption{Analysis of RMSNorm effect on SelFusion. We analyze tokens that are masked in both modes. (a) shows the probability distribution when both modes are correct, while (b) shows the case when both modes are incorrect. Easy mode uses 50\% masking ratio (MR), and hard mode uses 75\% MR.}
    \label{fig:selfusion_rmsnorm_effect}
\end{figure*}
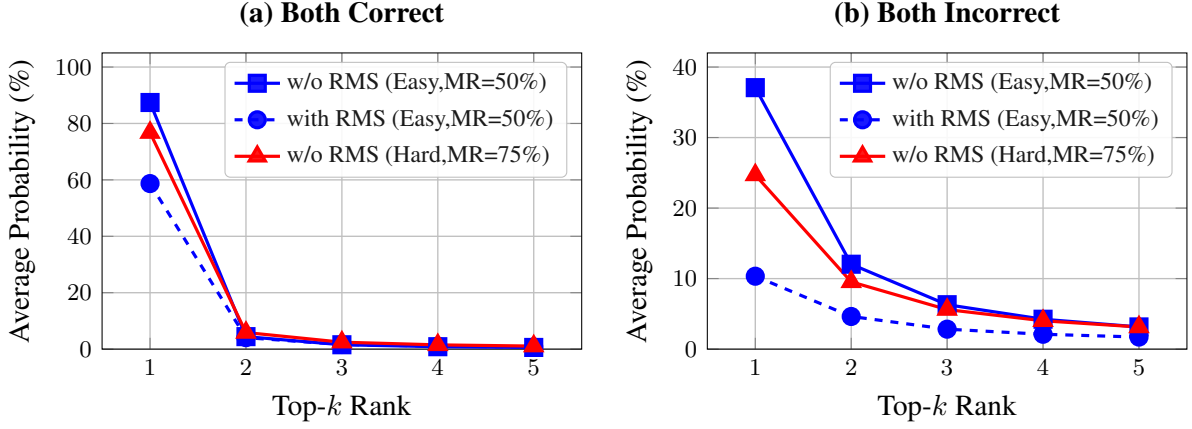

\subsection{Analysis}

\subsubsection{Training Efficiency}
We analyzed the training cost of SelFusion using TFLOPs. As shown in Table~\ref{tab:training_cost}, SelFusion required about 2$\times$ more per-iteration computation than SFT, but substantially less total computation than KD methods with a separately trained teacher. By removing the teacher-training stage, SelFusion reduced the overall training cost by roughly 2$\times$ while maintaining competitive performance. All results were measured under the same training configuration with 485 iterations.


\subsubsection{Effect of Bidirectional Distillation}
We further analyzed the effect of distillation direction by comparing easy$\rightarrow$hard, hard$\rightarrow$easy, and bidirectional KD. Although the easy mode is expected to be a stronger teacher due to its richer visible context, Table~\ref{tab:selfusion_ablation} shows that hard$\rightarrow$easy outperformed easy$\rightarrow$hard on several benchmarks, while neither unidirectional direction matched bidirectional distillation. As illustrated in Figure~\ref{fig:method}, SelFusion dynamically switches the token-level distillation direction between the easy and hard modes based on accuracy and confidence, allowing each mode to guide the other when it produces more confident predictions. Figure~\ref{fig:accuracy_comparison} provides quantitative support for this behavior. Under identical settings, easy$\rightarrow$hard distillation (denoted as Naive KD) resulted in an approximately 1\% drop in masked token accuracy for both modes during training.
Consistent with this, Table~\ref{tab:selfusion_ablation} shows that removing bidirectional KD causes substantial performance drops across all benchmarks, highlighting the importance of dynamic distillation target selection in self-distillation. 

\subsubsection{Token-level Probability Comparison of Easy and Hard Modes}

\label{sec:easy_hard_reliability}

To verify whether the easy mode assigns higher probability to ground-truth tokens than the hard mode, we analyze the model outputs under a controlled masking setup. Figure~\ref{fig:easy_hard_token_prob_by_step} presents the mean probability assigned to ground-truth tokens across training steps. Specifically, we evaluated step-wise checkpoints of SelFusion by fixing the hard mode masking ratio to 60\% and using a reduced ratio of 30\% for the easy mode, where the easy mode mask was constructed as a subset of the hard mode mask. We then measured the mean probability assigned to ground-truth tokens over this shared masked subset. The easy mode consistently assigned higher probability to the correct token than the hard mode, with the gap increasing from approximately 6\% to about 10\% over training. This observation supported our design intuition of treating the easy mode as the teacher and the hard mode as the student. Importantly, the easy mode maintains an output distribution that is more closely aligned with the student model, thereby facilitating effective knowledge transfer in line with prior student-friendly KD principles \citep{gu2025minillmknowledgedistillationlarge, kim2024promptkddistillingstudentfriendlyknowledge}.

\subsubsection{Logit Calibration by RMSNorm}
\label{sec:RMSNorm}
We applied RMSNorm-based logit calibration to mitigate overconfident predictions from the easy mode. As shown in Figure~\ref{fig:selfusion_rmsnorm_effect}, RMSNorm selectively calibrated confidence depending on prediction correctness. When both modes were correct, RMSNorm moderately reduced the top-1 probability from approximately around 90\% to about 60\%. More importantly, when both modes were incorrect, RMSNorm substantially suppressed the overconfident probability from around 40\% to about 10\%, approximately 75\% reduction. This stronger calibration on incorrect predictions was crucial, as it prevented learning from unreliable signals. Table~\ref{tab:selfusion_ablation} further confirmed that removing RMSNorm resulted in consistent performance drops, indicating its importance in suppressing overly confident incorrect predictions during distillation.

\newcommand{\twoscore}[2]{%
\makecell[c]{\strut #1\\[2pt]\makebox[\linewidth][r]{\scriptsize(#2)}\strut}%
}

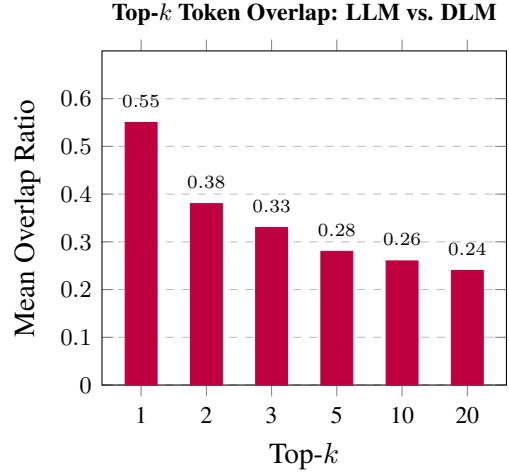
\begin{figure}[t]
    \raggedright
    \begin{tikzpicture}
        \begin{axis}[
            width=0.9\linewidth,
            height=6cm,
            title={\textbf{Top-$k$ Token Overlap: LLM vs. DLM}},
            title style={font=\small},
            xlabel={Top-$k$},
            ylabel={Mean Overlap Ratio},
            ymin=0, ymax=0.7,
            ytick={0, 0.1, 0.2, 0.3, 0.4, 0.5, 0.6},
            ymajorgrids=true,
            xmajorgrids=false,
            grid style={dashed, line width=0.4pt, draw=gray!50},
            label style={font=\normalsize},
            tick label style={font=\small},
            symbolic x coords={1,2,3,5,10,20},
            xtick=data,
            nodes near coords,
            every node near coord/.append style={
                font=\scriptsize,
                yshift=2pt,
            },
            ybar,
            bar width=12pt,
            enlarge x limits=0.12,
        ]
            \addplot[
                fill=purple,
                draw=purple,
            ] coordinates {
                (1, 0.55)
                (2, 0.38)
                (3, 0.33)
                (5, 0.28)
                (10, 0.26)
                (20, 0.24)
            };
            
        \end{axis}
    \end{tikzpicture}
    \caption{Mean top-$k$ token overlap between LLM and DLM fine-tuned on the Dolly dataset. For each position, overlap is computed as $|S_k^{\mathrm{LLM}} \cap S_k^{\mathrm{DLM}}| / k$, where $S_k$ denotes the set of top-$k$ predicted tokens.}
    \label{fig:topk_overlap}
\end{figure}

\begin{table*}[t]
\centering
{\fontsize{9}{12}\selectfont
\renewcommand{\arraystretch}{1.3}
\begin{tabular}{
>{\centering\arraybackslash}m{2.2cm}|
>{\centering\arraybackslash}m{3.0cm}|
>{\centering\arraybackslash}m{0.9cm}|
>{\centering\arraybackslash}m{1.25cm}|
>{\centering\arraybackslash}m{1.25cm}|
>{\centering\arraybackslash}m{1.25cm}|
>{\centering\arraybackslash}m{1.25cm}|
>{\centering\arraybackslash}m{1.25cm}
}
\noalign{\hrule height 1.5pt}
\textbf{Method} & \textbf{KD} & \textbf{Steps} &
\cellcolor{softgray}\textbf{Dolly} &
\cellcolor{softgray}\textbf{Self-inst} &
\cellcolor{softgray}\textbf{Vicuna} &
\cellcolor{softgray}\textbf{Sinst} &
\cellcolor{softgray}\textbf{Uinst} \\
\noalign{\hrule height 1.5pt}
\multirow{2}{*}{Logit-level KD}
& $\text{LLM}_{tea}$$\rightarrow$$\text{DLM}_{stu}$ & \multirow{5}{*}{32} 
&14.7424  & 8.9812 & 15.2372 & 15.6568 & 17.1595 \\
\cline{2-2}\cline{4-8}  
& $\text{DLM}_{tea}$$\rightarrow$$\text{DLM}_{stu}$ &
&18.8043  & 11.6876 & 16.0992 &26.8568  & 26.1251 \\
\cline{1-2}\cline{4-8} 
\multirow{2}{*}{Seq-level KD}
& $\text{LLM}_{tea}$$\rightarrow$$\text{DLM}_{stu}$ &
& 20.1426 &11.7443 & 16.6505 & 24.1393 &  25.1063\\
\cline{2-2}\cline{4-8}
& $\text{DLM}_{tea}$$\rightarrow$$\text{DLM}_{stu}$ &
& 15.9906 & 9.8436 & \textbf{17.2801} & 16.9256 & 18.6831  \\
\cline{1-2}\cline{4-8}
\textbf{Ours, SelFusion}
& $\text{DLM}_{stu}$$\leftrightarrow$$\text{DLM}_{stu}$ &
& \textbf{21.7828} & \textbf{12.4917} & 17.1310 & \textbf{27.1185} & \textbf{28.1247} \\
\noalign{\hrule height 1.5pt}
\end{tabular}
}
\caption{Ablation on larger diffusion steps. We evaluate DLMs with 32 diffusion steps to assess generalization beyond the main 8 and 16 step settings. Bold indicates the best result.
}
\label{tab:results_for_larger_steps}
\end{table*}

\subsubsection{Distribution Mismatch between LLMs and DLMs}
\label{sec:distribution_mismatch}
The primary challenge of LLM-to-DLM distillation stemmed from logit distribution mismatch caused by different generation mechanisms. We categorized this mismatch into two types: (1) top-$k$ logit scale mismatch and (2) top-$k$ token mismatch. As shown in Figure~\ref{fig:maksing-level}, LLMs and DLMs exhibited different probability scales: LLMs showed peaked distributions dominated by the top-1 token, whereas DLMs exhibited flatter distributions due to parallel generation. Figure~\ref{fig:topk_overlap} also showed limited top-$k$ token overlap, with top-1 overlap at about 60\% and decreasing as $k$ increased. These mismatches hindered direct logit-level distillation from LLMs to DLMs, which SelFusion addressed via the inherent generation mechanism of DLMs.

\begin{table}
\centering
\begin{tabular}{c|c|ccc}
\toprule[1.2pt]
  & \textbf{Step} & \textbf{Latency (ms)} & \textbf{Speed-up} \\
\midrule
\textbf{AR} 
 & -- & 2240 & 1.00$\times$ \\
\midrule
\multirow{7}{*}{\textbf{DLM}}
 & 1  & 61.4   & 36.5$\times$ \\
 & 2  & 109.7  & 20.4$\times$ \\
 & 4  & 199.1  & 11.3$\times$ \\
 & 8  & 378.4  & 5.9$\times$  \\
 & 16 & 745.4  & 3.0$\times$  \\
 & 32 & 1479.0 & 1.5$\times$  \\
 & 64 & 2952.2 & 0.76$\times$ \\
\bottomrule[1.2pt]
\end{tabular}
\caption{Per-sample inference latency and speed-up for AR and DLM with varying diffusion steps, measured on the Dolly validation set.}
\label{tab:time_comparison}
\end{table}

\subsubsection{Generalization on Larger Steps}
\label{sec:ablation}
We further evaluated its generalization to larger diffusion step settings. Beyond the standard 8 and 16 steps, we additionally evaluated 32-step inference. As shown in Table~\ref{tab:results_for_larger_steps}, SelFusion generally outperforms baseline distillation methods under the 32-step setting, achieving the best average performance across benchmarks.

\subsection{Time Comparison of DLM and LLM}

We compared the inference speed of DLMs and AR language models (LLMs).
As shown in Table~\ref{tab:time_comparison}, DLMs achieved significantly lower inference latency than LLMs.
This advantage stemmed from the parallel token generation of DLMs, whereas LLMs generated tokens sequentially.
As a result, even with 32 diffusion steps, DLMs achieved approximately a 1.5$\times$ speedup over LLMs in practice.

\section{Conclusions}

In this paper, we propose SelFusion, a novel self-distillation framework that enables effective logit-level KD for DLMs.
By leveraging different masking ratios with simultaneous forward passes, we decompose the model into two modes.
This provides a more suitable distribution for learning, enabling effective training within a single model.
Experimental results show that SelFusion outperforms conventional KD methods without relying on external teacher models.

\section*{Limitations}
Despite SelFusion's consistent performance gains, several limitations remain. First, the current availability of DLM backbones is limited, constraining validation across a broader set of architectures. 
Second, our evaluation is limited to instruction tuning in English, leaving broader domains and multilingual settings for future work. Despite these limitations, SelFusion offers a practical advantage as a distillation framework that does not rely on external teacher models, including LLMs or DLMs.

\section*{Ethical Considerations}
This work does not raise any ethical concerns.
\section*{Acknowledgments}
This work was supported by the Institute of Information \& Communications Technology Planning \& Evaluation (IITP) grant funded by the Korea government (MSIT) (RS-2025-02653113, High-Performance Research AI Computing Infrastructure Support at the 2 PFLOPS Scale). This work was also supported by the Institute of Information \& Communications Technology Planning \& Evaluation (IITP) grant funded by the Korea government (MSIT) (RS-2021-II211341, Artificial Intelligence Graduate School Program (Chung-Ang University)), and the National Research Foundation of Korea (NRF) grant funded by the Korea government (MSIT) (RS-2025-00515722).
\bibliography{custom}
\appendix
\section{Appendix}
\label{sec:appendix}
\subsection{Training Configuration}
We present the detailed training configurations used in our experiments in Table~\ref{tab:hparams_all}. We determined the optimal learning rates through grid search, taking into account both model scale and architectural differences. For the 1.4B teacher models, including both AR and DLM architectures, we explored a wider range of learning rates, $\{1\text{e-}5, 5\text{e-}5, 1\text{e-}4, 2\text{e-}4\}$, due to their distinct architectural characteristics. For the 0.472B DLM target models, we conducted grid search over $\{5\text{e-}5, 1\text{e-}4, 2\text{e-}4\}$. Following the same procedure, we also selected the learning rate for SelFusion via grid search and used $5\text{e-}5$ in the final configuration. The table also reports hyperparameters shared across all experimental settings. For evaluation, we used three random seeds, 10, 20, and 30, and report the average over these runs.

\begin{table}[h]
\centering
\fontsize{10}{12}\selectfont
\setlength{\tabcolsep}{4pt}
\renewcommand{\arraystretch}{1.15}
\begin{tabular}{@{} l p{0.60\columnwidth} r @{}}
\toprule
\textbf{Stage} & \textbf{Setting} & \textbf{Value} \\
\midrule
\multirow{4}{*}{SFT}
& DLM (472M)   & $5\times10^{-5}$ \\
& DLM (1.476B) & $1\times10^{-5}$ \\
& LLM (472M)   & $2\times10^{-4}$ \\
& LLM (1.476B) & $1\times10^{-4}$ \\
\midrule
\multirow{4}{*}{KD}
& DLM$\rightarrow$DLM (logit KD) & $5\times10^{-5}$ \\
& DLM$\rightarrow$DLM (seq KD)   & $5\times10^{-5}$ \\
& LLM$\rightarrow$DLM (logit KD) & $5\times10^{-5}$ \\
& LLM$\rightarrow$DLM (seq KD)   & $5\times10^{-5}$ \\
\midrule
\multicolumn{2}{@{}l}{\textbf{Shared hyperparameters}} & \textbf{Value} \\
\midrule
\multicolumn{2}{@{}l}{\# devices} & 4 \\
\multicolumn{2}{@{}l}{Global batch size} & 512 \\
\multicolumn{2}{@{}l}{Max tokens} & 256 \\
\multicolumn{2}{@{}l}{Epoch (final)} & 20 \\
\multicolumn{2}{@{}l}{LR decay} & enabled \\
\multicolumn{2}{@{}l}{Warmup ratio} & 0.05 \\
\multicolumn{2}{@{}l}{Min LR} & LR$/10$ \\
\multicolumn{2}{@{}l}{Weight decay} & 0.1 \\
\multicolumn{2}{@{}l}{Adam betas} & $(0.9,\,0.95)$ \\
\multicolumn{2}{@{}l}{Grad clip} & 1.0 \\
\multicolumn{2}{@{}l}{Seed} & 3407 \\
\bottomrule
\end{tabular}
\vspace{-2mm}
\caption{Selected hyperparameters from grid search and shared training settings.}
\label{tab:hparams_all}
\vspace{-2mm}
\end{table}

\subsection{Prompt Formatting for Instruction Tuning}
\label{sec:appendix_data_preprocess}

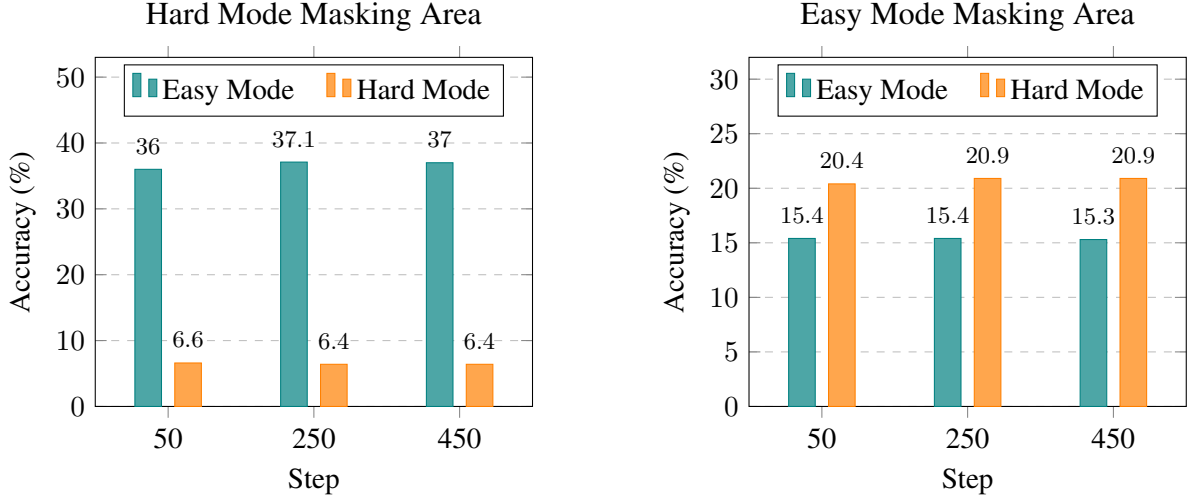
\begin{figure*}[t]
\centering
\begin{subfigure}[b]{0.46\textwidth}
\centering
\begin{tikzpicture}
\begin{axis}[
    width=\linewidth,
    height=6.2cm,
    title={Hard Mode Masking Area},
    xlabel={Step},
    ylabel={Accuracy (\%)},
    symbolic x coords={50,250,450},
    xtick=data,
    ymin=0, ymax=53,
    ytick={0,10,20,30,40,50},
    ymajorgrids=true,
    grid style={dashed, line width=0.4pt, draw=gray!50},
    ybar=5pt,
    bar width=10pt,
    enlarge x limits=0.25,
    tick label style={font=\normalsize},
    label style={font=\normalsize},
    title style={font=\large},
    nodes near coords,
    nodes near coords align={vertical},
    nodes near coords style={
        font=\small,
        yshift=2pt
    },
    legend style={
        at={(0.5,0.98)},
        anchor=north,
        legend columns=2,
        font=\normalsize,
        draw=black,
        fill=white,
        inner sep=3pt,
        /tikz/every even column/.append style={column sep=10pt}
    },
    legend cell align={left},
]
\addplot[fill=teal!70, draw=teal] coordinates {
    (50,36.0) (250,37.1) (450,37.0)
};
\addplot[fill=orange!70, draw=orange] coordinates {
    (50,6.6) (250,6.4) (450,6.4)
};
\legend{Easy Mode, Hard Mode}
\end{axis}
\end{tikzpicture}
\end{subfigure}
\hfill
\begin{subfigure}[b]{0.46\textwidth}
\centering
\begin{tikzpicture}
\begin{axis}[
    width=\linewidth,
    height=6.2cm,
    title={Easy Mode Masking Area},
    xlabel={Step},
    ylabel={Accuracy (\%)},
    symbolic x coords={50,250,450},
    xtick=data,
    ymin=0, ymax=32,
    ytick={0,5,10,15,20,25,30},
    ymajorgrids=true,
    grid style={dashed, line width=0.4pt, draw=gray!50},
    ybar=5pt,
    bar width=10pt,
    enlarge x limits=0.25,
    tick label style={font=\normalsize},
    label style={font=\normalsize},
    title style={font=\large},
    nodes near coords,
    nodes near coords align={vertical},
    nodes near coords style={
        font=\small,
        yshift=2pt
    },
    legend style={
        at={(0.5,0.98)},
        anchor=north,
        legend columns=2,
        font=\normalsize,
        draw=black,
        fill=white,
        inner sep=3pt,
        /tikz/every even column/.append style={column sep=10pt}
    },
    legend cell align={left},
]
\addplot[fill=teal!70, draw=teal] coordinates {
    (50,15.4) (250,15.4) (450,15.3)
};
\addplot[fill=orange!70, draw=orange] coordinates {
    (50,20.4) (250,20.9) (450,20.9)
};
\legend{Easy Mode, Hard Mode}
\end{axis}
\end{tikzpicture}
\end{subfigure}
\caption{Masked-token prediction accuracy comparison between the easy and hard modes. The left plot evaluates both modes on hard mode masked positions, and the right plot evaluates both modes on easy mode masked positions. Note that a token masked in one mode may remain visible in the other mode. Thus, for a token masked in one mode, the counterpart mode may observe the ground truth token at that position.}
\label{fig:left_right_bar}
\end{figure*}

We standardized the instruction tuning data by converting each example into a Dolly style prompt template. Specifically, when an example contains an \textbf{input field}, we construct the prompt as follows:

\begin{tcolorbox}[colback=gray!5, colframe=gray!50, arc=2pt, boxrule=0.5pt, left=5pt, right=5pt, top=5pt, bottom=5pt]
\small\ttfamily
Below is an instruction that describes a task, paired with an input that provides further context. Write a response that appropriately completes the request. \\

\#\#\# Instruction: \\
\{instruction\} \\

\#\#\# Input: \\
\{context\} \\

\#\#\# Response:
\end{tcolorbox}
When the \textbf{input field is absent}, the following template is used:
\begin{tcolorbox}[colback=gray!5, colframe=gray!50, arc=2pt, boxrule=0.5pt, left=5pt, right=5pt, top=5pt, bottom=5pt]
\small\ttfamily
Below is an instruction that describes a task. Write a response that appropriately completes the request. \\

\#\#\# Instruction: \\
\{instruction\} \\

\#\#\# Response:
\end{tcolorbox}
All datasets used in our experiments are publicly available from the \textbf{MiniLLM} data release: \\
\url{https://github.com/microsoft/LMOps/tree/main/minillm}

\subsection{Masking Strategy for Easy and Hard Modes}
\label{sec:masking_config}

We specify the masking configuration for the easy and hard modes following the standard DLM noising procedure.
Given an input sequence $\mathbf{x}\in\{0,\dots,V-1\}^{L}$, we sample a noise level $t\sim\mathcal{U}(0,1)$ for each example and convert it into a token masking probability
\begin{equation}
p_{\text{mask}}(t) = (1-\epsilon)\,t + \epsilon,
\end{equation}
where $\epsilon$ is a small constant to avoid degenerate masking.
We then independently mask each position $i$ with probability $p_{\text{mask}}(t)$ and replace masked tokens with a dedicated mask token (implemented by using the vocabulary index $V$):
\begin{equation}
\tilde{x}_i =
\begin{cases}
\texttt{[MASK]} & \text{with prob. } p_{\text{mask}}(t),\\
x_i & \text{otherwise}.
\end{cases}
\end{equation}
To construct paired easy and hard modes within a single training step, we first sample the hard mode noise level $t_{\text{hard}}\sim\mathcal{U}(0,1)$.
We then sample the easy mode noise level conditioned on it as
\begin{equation}
t_{\text{easy}} \sim \mathcal{U}(0,t_{\text{hard}}),
\end{equation}
which ensures $t_{\text{easy}} \le t_{\text{hard}}$ and thus $p_{\text{mask}}(t_{\text{easy}})\le p_{\text{mask}}(t_{\text{hard}})$.
Accordingly, the easy mode observes more visible context (lower masking), while the hard mode operates under reduced visibility (higher masking).
Although the easy mode masking ratio is determined by conditioning on the hard mode noise level, the specific masked token positions are sampled independently for the two modes.

\begin{table*}[t]
\centering
{\fontsize{9}{12}\selectfont
\renewcommand{\arraystretch}{1.3}
\begin{tabular}{
>{\centering\arraybackslash}m{3.0cm}|
>{\centering\arraybackslash}m{3.5cm}|
>{\centering\arraybackslash}m{2.0cm}|
>{\centering\arraybackslash}m{2.0cm}
}
\noalign{\hrule height 1.5pt}
\textbf{Method} & \textbf{KD} &
\cellcolor{softgray}\textbf{8 Steps} &
\cellcolor{softgray}\textbf{16 Steps} \\
\noalign{\hrule height 1.5pt}
SFT & --- & 27.9175 & 29.3172 \\
\hline
\multirow{2}{*}{Seq-level KD}
& $\text{LLM}_{tea}\rightarrow\text{DLM}_{stu}$ & 27.7835 & 29.2686 \\
\cline{2-4}
& $\text{DLM}_{tea}\rightarrow\text{DLM}_{stu}$ & 21.9711 & 23.4053 \\
\hline
\multirow{2}{*}{Logit-level KD}
& $\text{LLM}_{tea}\rightarrow\text{DLM}_{stu}$ & 26.3727 & 26.6891 \\
\cline{2-4}
& $\text{DLM}_{tea}\rightarrow\text{DLM}_{stu}$ & 28.9661 & 30.0167 \\
\hline
\textbf{Ours, SelFusion}
& $\text{DLM}_{stu}\leftrightarrow\text{DLM}_{stu}$ & \textbf{29.3575} & \textbf{30.4928} \\
\noalign{\hrule height 1.5pt}
\end{tabular}
}
\caption{Results on SAMSum measured by ROUGE-L across different diffusion steps. SelFusion consistently outperforms strong baselines, with larger gains under the more efficient 8-step setting. Bold indicates the best result.}
\label{tab:samsum_results}
\end{table*}

\subsection{Token level Accuracy Comparison Details}
\label{sec:appendix_token_acc}
To examine whether bidirectional KD is activated during training, we analyze the probability that only one of the two modes correctly predicts a masked token. As shown in Figure~\ref{fig:left_right_bar}, we measure this probability on masked token positions for each mode. In the left plot, which evaluates hard mode masked tokens, we observe that only the hard mode predicts the correct token in approximately 6\% of cases, whereas the easy mode alone is correct in about 37\% of cases.
In contrast, in the right plot corresponding to easy mode masked tokens, the hard mode correctly predicts the token in around 20\% of cases, while the easy mode alone is correct in only about 15\% of cases.
These results indicate that even on its own masked positions, the easy mode is not always more accurate, and the hard mode can provide more accurate token predictions depending on the masking configuration.
This complementary behavior explains why SelFusion benefits from bidirectional KD, as distillation can dynamically proceed from the mode with higher token accuracy at each masked position.

\subsection{Generalization Across Tasks}
To evaluate the generality of SelFusion beyond instruction-following, we further tested it on the summarization benchmark SAMSum \citep{gliwa-etal-2019-samsum}. As shown in Table \ref{tab:samsum_results}, SelFusion outperformed strong baselines, including SFT and existing KD methods, in both the 8-step and 16-step settings. The improvement was more pronounced in the 8-step setting, indicating that SelFusion remained effective under tighter inference budgets.

\end{document}